\documentclass[11pt,a 4paper]{article}
\usepackage[final]{acl}

\usepackage{times,latexsym}
\usepackage{url}
\usepackage[T1]{fontenc}
\usepackage{multirow}
\usepackage{amsmath}
\usepackage{diagbox}
\usepackage{caption}
\usepackage{subcaption}
\usepackage{graphicx}
\usepackage{amssymb}
\usepackage{mathtools}
\usepackage{adjustbox}
\usepackage{balance}

\newcommand{\system}[1]{\textsc{#1}}
\newcommand{\data}[1]{\textsc{#1}}

\newcommand{\task}{UDA-ECPE\xspace}
\newcommand{\ourmethod}{\system{CD-SelfTrain}\xspace}
\newcommand{\ourmodel}{\system{CaRel-VAE}\xspace}

\newcommand{\enecpe}{\data{EN-ECPE}\xspace}
\newcommand{\checpe}{\data{CH-ECPE}\xspace}

\newcommand{\heart}{\ensuremath\heartsuit}
\newcommand{\diamondsmall}{\ensuremath\diamondsuit}
\newcommand{\club}{\ensuremath\clubsuit}

\usepackage{rotating}

\usepackage{xspace,mfirstuc,tabulary}

\usepackage{amsthm,amsmath,amsfonts,bm,xspace}
\usepackage{color}

\theoremstyle{plain}

\theoremstyle{definition}

\theoremstyle{remark}

\def\eqref#1{(\ref{#1})}

\def\1{\bm{1}}

\def\rmI{{\mathbf{I}}}

\def\rmX{{\mathbf{X}}}

\def\rmZ{{\mathbf{Z}}}

\def\vh{{\bm{h}}}

\def\vx{{\bm{x}}}

\def\vz{{\bm{z}}}

\def\mSigma{{\bm{\Sigma}}}

\DeclareMathAlphabet{\mathsfit}{\encodingdefault}{\sfdefault}{m}{sl}
\SetMathAlphabet{\mathsfit}{bold}{\encodingdefault}{\sfdefault}{bx}{n}

\def\gD{{\mathcal{D}}}

\def\gL{{\mathcal{L}}}

\def\gR{{\mathcal{R}}}

\def\gY{{\mathcal{Y}}}

\newcommand{\sigmoid}{\sigma}

\newcommand{\KL}{\mathbb{D}_{\mathrm{KL}}}
\newcommand{\diver}{\mathbb{D}}

\newcommand{\vMu}{\bm{\mu}}
\newcommand{\vSigma}{\bm{\sigma}}

\newcommand{\targettask}{ECPE\xspace}

\newif\iftaclinstructions
\taclinstructionsfalse 
\iftaclinstructions
\renewcommand{\confidential}{}
\renewcommand{\anonsubtext}{(No author info supplied here, for consistency with
TACL-submission anonymization requirements)}
\newcommand{\instr}
\fi

\title{Causal Discovery Inspired Unsupervised Domain Adaptation for Emotion-Cause Pair Extraction}

\author{Yuncheng Hua\textsuperscript{\rm \heart}\footnotemark[1], Yujin Huang\textsuperscript{\rm \heart}\footnotemark[1], Shuo Huang\textsuperscript{\rm \heart}, Tao Feng\textsuperscript{\rm \heart}, Lizhen Qu\textsuperscript{\rm \heart}\footnotemark[2], \\ \textbf{Chris Bain}\textsuperscript{\rm \club},  
\textbf{Richard Bassed}\textsuperscript{\rm \diamondsmall}, \textbf{Gholamreza Haffari}\textsuperscript{\rm \heart} \\
\textsuperscript{\rm \heart} Department of Data Science \& AI, Monash University, Australia\\
\textsuperscript{\rm \club} Department of Human Centred Computing, Monash University, Australia\\
\textsuperscript{\rm \diamondsmall} Victorian Institute of Forensic Medicine, Melbourne, Australia \\
\{devin.hua, yujin.huang, shuo.huang1, tao.feng\}@monash.edu, \\ \{lizhen.qu, chris.a.bain,  gholamreza.haffari\}@monash.edu, \\ Richard.Bassed@vifm.org\\ 
}

\date{}

\begin{document}
\maketitle

\renewcommand{\thefootnote}{\fnsymbol{footnote}}
\footnotetext[1]{Equal contribution.}
\footnotetext[2]{Corresponding author.}

\begin{abstract}
This paper tackles the task of emotion-cause pair extraction in the unsupervised domain adaptation setting.
The problem is challenging as the distributions of the events causing emotions in target domains are dramatically different than those in source domains, despite the distributions of emotional expressions between domains are overlapped. Inspired by causal discovery,
we propose a novel deep latent model in the variational autoencoder (VAE) framework, which not only captures the underlying latent structures of data but also utilizes the easily transferable knowledge of emotions as the bridge to link the distributions of events in different domains. 
To facilitate knowledge transfer across domains, we also propose a novel variational posterior regularization technique to disentangle the latent representations of emotions from those of events in order to mitigate the damage caused by the spurious correlations related to the events in source domains. Through extensive experiments, we demonstrate that our model outperforms the strongest baseline by approximately 11.05\% on a Chinese benchmark and 2.45\% on a English benchmark in terms of weighted-average F1 score. 
We have released our source code and the generated dataset publicly at: {\small\textsf{\url{https://github.com/tk1363704/CAREL-VAE}}}.
\end{abstract}
\section{Introduction}
\begin{figure}[hbt!]
    \centering
    \includegraphics[width=\linewidth]{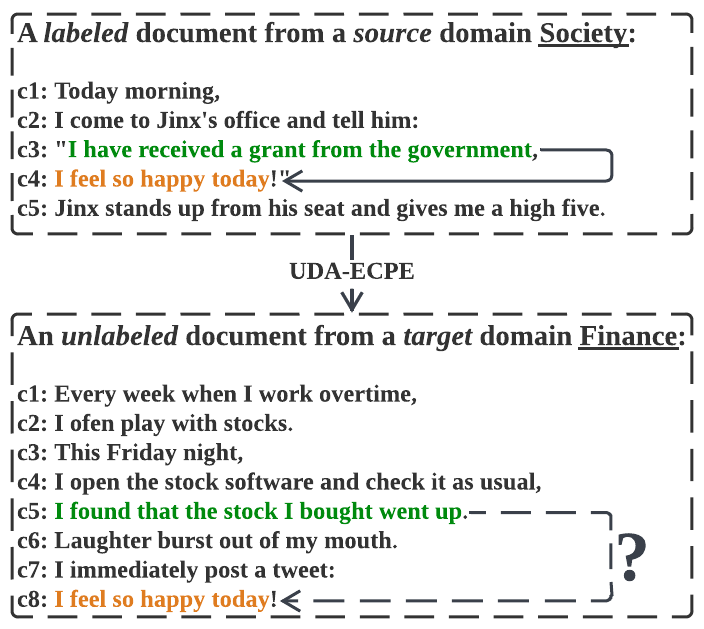}
    \caption{An illustrative example of the \task task. Orange and green highlights respectively denote emotion and cause clauses.}
    \label{fig: UDA_ECPE_task}
\end{figure}

Emotion-cause pair extraction (\targettask) aims to extract emotions and the events causing such emotions mentioned in a document~\cite{xia2019emotion}. The task has potential applications in a number of areas, such as affective computing, market analysis, and intelligent agents for customer support. However, there are only a small number of labeled training corpora available in a handful of domains. As shown in Fig. \ref{fig: UDA_ECPE_task}, in order to deploy \targettask models to target domains, where there are only unlabeled data, we focus on the unsupervised domain adaptation (UDA) for \targettask, coined \task, which is not explored before.

Multi-class or multi-label classification dominates in conventional UDA tasks. \task is more challenging because the events causing the same emotion are barely the same across domains, despite the knowledge of emotional expressions is easier to transfer across domains using the UDA methods~\cite{zad2021emotion}. For example, the reason for "I feel so happy today" can be "I have received a grant from the government" in the society domain and "I found that the stock I bought went up" in the finance domain. There are usually no explicit keywords such as "because" showing their causal relations. However, current UDA methods assume that there are small discrepancies between source and target distributions~\cite{zhao2019learning,kumar2020understanding}. We show in Sec. \ref{sec:results} that the state-of-the-art (SOTA) UDA methods indeed have limited capabilities to improve the performance of the SOTA \targettask models. 

It is a common practice to project texts into latent representations for improving language understanding~\cite{wang2019superglue}. 
Existing techniques disentangle different types of latent representations by applying regularization terms to enforce independence between the corresponding random variables~\cite{cheng2020improving}. However, the independence assumption \textit{contradicts} the fact that emotions and the events causing them are \textit{statically dependent}. 
Furthermore, Large Language Models (LLMs), as a powerful general-purpose natural language processing tool, should be well-suited for the \task task. 
However, in many related studies~\cite{DBLP:conf/naacl/GaoCH19,DBLP:journals/tmlr/ZecevicWDK23,kiciman2023causal,DBLP:conf/emnlp/RomanouMPLAB23,DBLP:conf/emnlp/JacoviCGG23,DBLP:conf/emnlp/GaoD0023,DBLP:conf/iclr/Jin0LPSMDS24,DBLP:journals/corr/abs-2407-19638}, researchers have found that LLMs are not particularly effective at solving causal discovery tasks.

To tackle the above challenges, we take the transferable knowledge of emotional expressions as the bridge between a source domain and a target domain. In a single domain, we identify causal relations between emotions and domain-specific events, which can be viewed as a causal discovery problem between the corresponding random variables. In the VAE framework~\cite{kingma2013auto}, we propose a \textit{novel} model, coined \ourmodel, to map inputs texts into latent emotion representations and latent event representations and detect their causal relations. Herein, we propose a \textit{novel} variational posterior regularizer to disentangle those representations by maximizing the divergences between the posteriors without assuming independence. In a target domain, we improve the self-training algorithm~\cite{chen2011co} for discovering domain-specific causal relations, referred to as \ourmethod. Instead of incrementally updating a training set, we improve the original algorithm by producing a new pseudo-labeled training set in each epoch. As a result, our method outperforms the SOTA \targettask models trained with the SOTA UDA methods by a wide margin. 

To sum up, our contributions are the following:
\begin{itemize}
    \item We propose a \textit{novel} causal discovery inspired UDA method, coined \ourmethod, and a \textit{new} model, coined \ourmodel, for the \targettask task in the unexplored UDA setting.
    \item We propose a novel disentanglement regularization term on variational Posteriors so that it does not enforce independence between emotions and the events causing them.
    \item Our approach achieves superior performance in terms of weighted-average F1 over the strongest baseline by approximately 11.05\% on a Chinese benchmark and 2.45\% on a English benchmark. Even if that baseline is trained with the SOTA UDA method, our method still achieves the best.
\end{itemize}
\section{Challenges in \task}
\label{sec:problem_analysis}
The task \targettask is concerned with recognizing causal relations between the events causing emotions and the corresponding emotional expressions mentioned in a document. All prior studies on the \targettask task employ a (deep) learning-based classifier to detect mentions of causal relations based on an input text. They often choose an input text that mentions an event and an emotional expression. Then those classifiers determine whether the event causes the emotional expression by investigating if i) the event and the emotional expression are correlated and ii) there is a linguistic pattern indicating their relation is causal, e.g. using a key phrase ``leads to''.

Formally, given an input text $\vx$, we extract an event embedding $\vz^c$ and an emotion embedding $\vz^e$, which are the values sampled from the corresponding latent random variable vectors $\rmZ^c$ and $\rmZ^e$. In a source domain, a model learns a distribution $\sum_{\rmZ^c, \rmZ^e}p(Y | \rmZ^c, \rmZ^e, \vx)p(\rmZ^c, \rmZ^e| \vx)$, where $Y$ denotes a binary random variable indicating if there is a causal relation between $\rmZ^c$ and $\rmZ^e$. The key challenge is that both $p(Y | \rmZ^c, \rmZ^e, \vx)$ and $p(\rmZ^c, \rmZ^e| \vx)$ are significantly different in target domains. Although prior studies show that $p(\rmZ^e| \vx)$ can be easily transferred from source domains to target domains~\cite{DBLP:journals/tmlr/WangLCWCK22}, the correlations between $\rmZ^c$ and $\rmZ^e$ are almost not transferable, because $p(\rmZ^c)$ are dramatically different between domains. Therefore, when adapting a model trained in a source domain to a target domain, the model needs to \textit{forget} the correlations between emotions and events from the source domain, followed by learning new correlations in the target domain.

To provide an intuitive understanding of the above challenges in the UDA setting, we visualize the clause embeddings, namely $p(\rmZ^c)$, for ground-truth emotion and emotion causes respectively on \checpe and \enecpe, and compare them with the sentence embeddings for a widely used domain adaptation corpus Amazon Reviews~\cite{blitzer2007biographies} using t-SNE. As the original \checpe are not partitioned based on domains, we manually assign each data point in the corpus with the corresponding domain label. Further details are provided in Sec. \ref{exp_setup}.

\begin{figure*}[hbt!]
     \centering
     \begin{subfigure}[b]{0.31\textwidth}
         \centering
         \includegraphics[width=\textwidth]{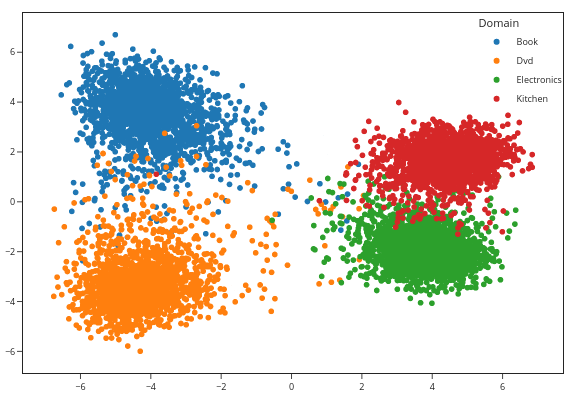}
         \caption{Amazon sentiment reviews}
         \label{fig: amazon}
     \end{subfigure}
     \hfill
     \begin{subfigure}[b]{0.31\textwidth}
         \centering
         \includegraphics[width=\textwidth]{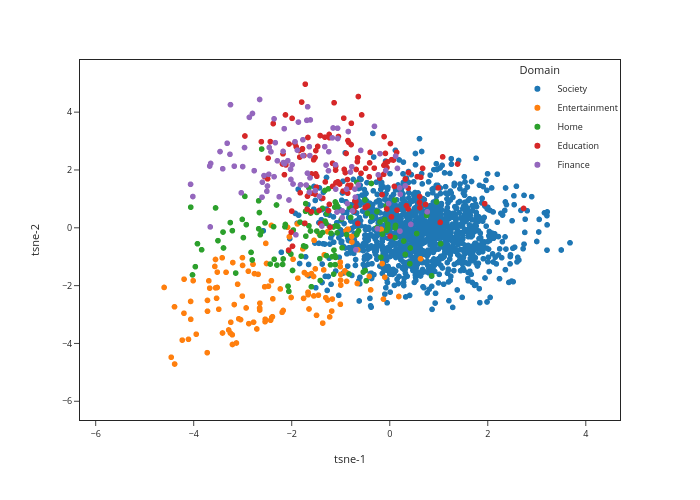}
         \caption{Chinese emotion clauses}
         \label{fig: chi_emotion}
     \end{subfigure}
     \hfill
     \begin{subfigure}[b]{0.31\textwidth}
         \centering
         \includegraphics[width=\textwidth]{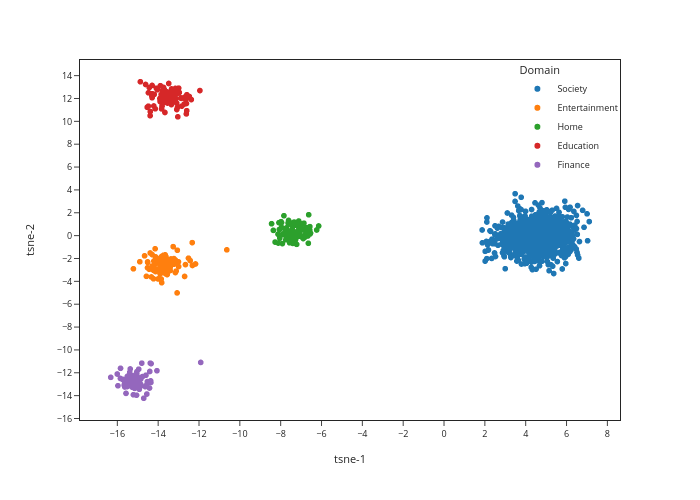}
         \caption{Chinese emotion cause clauses}
         \label{fig: chi_cause}
     \end{subfigure}
     \hfill

     \caption{The t-SNE visualizations of the sentence embeddings from Amazon Reviews multi-domain sentiment corpus and the clause embeddings from the Chinese \task corpora.}
     \label{fig:CD-ECPE_challenges}
\end{figure*}

As shown in Figure~\ref{fig:CD-ECPE_challenges}, the data points of Chinese emotion clauses from various \checpe's domains are strongly overlapped, the domain divergences are far smaller than those of the embeddings of the emotion causes. It is thus challenging for existing UDA methods, which work only in the cases that the distribution shift from a source domain to a target domain is small, as illustrated in Fig.\ref{fig: amazon}~\cite{zhao2019learning,kumar2020understanding}.

In addition, we employ two different datasets as different domains for English.
The English corpora similar tendency can be found in \ref{appedix:sec1}.
As shown in Fig.\ref{fig:EN-ECPE_challenges}, regardless if a clause mentions an emotion or an emotion cause, there is a very clear boundary between the two domains. Their domain differences are largely caused by the differences between the two datasets.

\section{Methodology}
The \task task is concerned with identifying causal relations between mentions of events and emotional expressions in target domains, which do not have labeled data. In the source domain, there is a set of labeled documents $\gD^s = \{(\rmX^{s}_{1}, \gR^{s}_{1}), (\rmX^{s}_{2}, \gR^{s}_{2}), ..., (\rmX^{s}_{n}, \gR^{s}_{n}) \}$. Each document $\rmX^{s}_k$ consists of a sequence of clauses $(\vx_1,\vx_2, ..., \vx_{d})$ and is annotated with a set of labeled emotion-cause pairs $\gR^s_k=\{(y^r_{ij},y^c_i, y^e_j)\}_{i,j}$, where $y^r_{ij}$ is a binary label indicating if $\vx_i$ is an event mention causing an emotion expressed in $\vx_j$, $y_i^c$ denotes whether $\vx_i$ is an event or not, and $y^e_j \in \gY^e$ denotes the category of the emotion. In this work, we consider the widely used six basic emotion categories: happiness, sadness, fear, disgust, anger, and surprise. Then the task is to identify a set of such causal relations and emotion categories $\gR^t_k=\{(y^r_{ij}, y^e_j)\}_{i,j}$ from each unlabeled document $k$ in target domains. In contrast, the prior studies~\cite{xia2019emotion} assume the training and test distributions are identical and emotional expressions are not categorized. Hence, our setting is more difficult and practical by considering emotion categories and distribution discrepancies between domains.    

\paragraph{\ourmodel Overview.} Denoted by $\rmZ^e$ and $\rmZ^c$ the latent random variable vectors for emotion and event respectively, we adopt the VAE framework to learn the latent distribution $p(y_{ij}^r, y^e, y^c, \rmX_{ij}, \rmZ^e, \rmZ^c)$ for a pair of clauses $\rmX_{ij} = (\vx_i, \vx_j)$, which is factorized into 
\begin{small}
\begin{displaymath}
\overbrace{p(y_{ij}^r| \rmZ^e, \rmZ^c) p(y^e| \rmZ^e)p(y^c | \rmZ^c)}^{\text{task-specific}} \overbrace{ p(\rmX_{ij}| \rmZ^e, \rmZ^c) p(\rmZ^e)p(\rmZ^c)}^{\text{standard VAE}}
\end{displaymath}
\end{small}
In addition to the standard components of VAE, such as the decoder $p(\rmX_{ij}| \rmZ^e, \rmZ^c)$, we include task-specific predictors: an emotion classifier $p(y^e| \rmZ^e)$, an emotion-cause relation classifier $p(y_{ij}^r| \rmZ^e, \rmZ^c)$, and an event predictor $p(y^c | \rmZ^c)$. 

To approximate the true distribution, we consider a factorized variational distribution $q(\rmZ^e, \rmZ^c| \rmX_{ij}) = q(\rmZ^e | \rmX_{ij}) q(\rmZ^c | \rmX_{ij})$, which correspond to an emotion encoder and an event encoder respectively. Then the variational lower bound (ELBO) takes the following form:
\begin{small}
    \begin{equation}\nonumber
\begin{aligned}
    &\mathbb{E}_{q(\rmZ^e, \rmZ^c| \rmX_{ij})} \log \big[p(\rmX_{ij}|  \rmZ^e, \rmZ^c) p(y_{ij}^r| \rmZ^e, \rmZ^c) \\&p(y^e| \rmZ^e)p(y^c | \rmZ^c) \big] - \KL(q(\rmZ^e | \rmX_{ij} \| p(\rmZ^e)) \\
    & - \KL(q(\rmZ^c | \rmX_{ij} \| p(\rmZ^c))
\end{aligned}
\end{equation}
\end{small}

\paragraph{Disentanglement.} In target domains, it is not desirable that the latent representation of an emotion is mixed with event information, which makes transfer of the knowledge about emotions across domains difficult, because events in target domains are not directly related to those in source domains. Therefore, we need to disentangle latent emotion representations from latent event representations for improving compositional generalization~\cite{russin2019compositional} without making the independence assumption.

In light of the above analysis, we propose a variational posterior regularization technique. The key idea is to regularize the model in the way that the dense regions of $q(\rmZ^e | \rmX_{ij})$ associate with only emotions, while those of $q(\rmZ^c | \rmX_{ij})$ associate with only events. The classifiers for $p(y^e| \rmZ^e)$ and $p(y^c| \rmZ^c)$ are in general smooth such that they consistently predict only one label in a dense region. If there is little overlap between the dense regions of $q(\rmZ^e | \rmX_{ij})$ and those of $q(\rmZ^c | \rmX_{ij})$, a dense region from either distribution is expected to associated with either an emotion category or a type of events estimated by one of the classifiers, under the maximum likelihood principle. In another word, we only need to add a regularizer to minimize the overlap between $q(\rmZ^e | \rmX_{ij})$ and $q(\rmZ^c | \rmX_{ij})$ such that their divergence is high.

In theory, the corresponding divergence measures $\KL(q(\rmZ^e | \rmX_{ij}) \| q(\rmZ^c | \rmX_{ij}))$ should not assume absolute continuity~\cite{royden1988real}, which requires that $q(Z^e_i | \rmX_{ij}) > 0$ for every $q(Z^c_i | \rmX_{ij}) > 0$, vice versa. In reality, a random variable $Z_i^e$ may have high probability in the region where a $Z_j^c$ has zero probability. To tackle this, we choose Bhattacharyya distance~\cite{bhattacharyya1946measure} and maximum mean discrepancy (MMD)~\cite{gretton2012kernel} respectively as a regularizer. Each of them has its own strength. More details are covered in Sec. \ref{sec:training}.

\begin{figure*}[hbt!]
\centering
\includegraphics[width=\linewidth]{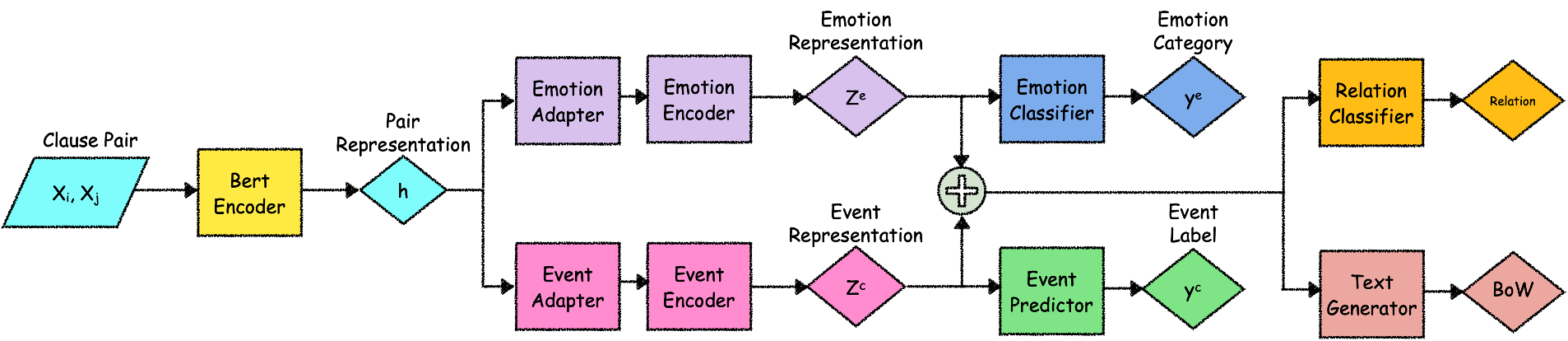}
\caption{The architecture of our model \ourmodel.}
\label{fig:mod_overview}
\end{figure*}
\subsection{Model Details}
\label{sec:model}

\paragraph{\ourmodel Model.} As illustrated in Fig.~\ref{fig:mod_overview}, our model is composed of an inference module, a text generator, task-specific predictors and priors.  

\emph{Inference Module.} The inference module consists of a pre-trained BERT~\cite{devlin2018bert} encoder, an emotion encoder and an event predictor. Given a pair of clauses $(\vx_i, \vx_j)$, we construct inputs following the common practice that inserts an $[SEP]$ token between the two clauses and prepends the sequence with a $[CLS]$ token. We take the hidden representation $\vh$ of $[CLS]$ as the output of the BERT encoder.

To distinguish the representation of the event and emotion variables, we employ two adapters to produce different embedding respectively. We initialize two vectors $\boldsymbol{a}_e$ and $\boldsymbol{a}_c$ for emotion and event respectively, and treat them as the queries while view  $\vh$ as key and value. We therefore synthesize the new emotion and event representations $\vh_e$ and $\vh_c$ by computing the sparsemax attention while using $\boldsymbol{a}_e$ and $\boldsymbol{a}_c$ as queries respectively~\cite{DBLP:conf/icml/MartinsA16}.

The variational distribution $q(\rmZ^e, \rmZ^c | \rmX_{ij})$ are realized as simple factorized Gaussians, which correpond to an emotion encoder $q(\rmZ^e | \vh_e)$ and an event predictor $q(\rmZ^c | \vh_c)$ on top of the hidden representations $\vh_e$ and $\vh_c$ respectively. Each encoder is implemented as a multilayer perceptrons (MLPs) after applying the reparameterization trick.

\begin{equation}
\begin{aligned}
\label{eq:emo_cau_encoders}
& \boldsymbol{\mu}^{e}, \log\vSigma^{e} = \text{MLP}(\vh_e; \boldsymbol{\theta}_e) \\
& \boldsymbol{\mu}^{c},\log\vSigma^{c} = \text{MLP}(\vh_c; \boldsymbol{\theta}_c) \\
& \vz^e = \boldsymbol{\mu}^{e} + \boldsymbol{\sigma}^{e}\odot \boldsymbol{\epsilon}, \boldsymbol{\epsilon}\sim\mathcal{N}(\mathbf{0}, \mathbf{I}) \\
& \vz^c = \boldsymbol{\mu}^{c} + \boldsymbol{\sigma}^{c}\odot \boldsymbol{\epsilon}, \boldsymbol{\epsilon}\sim\mathcal{N}(\mathbf{0}, \mathbf{I})
\end{aligned}
\end{equation}
where $\boldsymbol{\theta}_e$ and $\boldsymbol{\theta}_c$ are the parameters of the emotion and event encoders respectively, $\boldsymbol{\mu}^{e}$, $\boldsymbol{\sigma}^{e}$ and $\boldsymbol{\mu}^{c}$, $\boldsymbol{\sigma}^{c}$ denote the means and standard deviations of the corresponding Gaussian distributions, $\boldsymbol{\epsilon}$ denotes independent Gaussian noises, $\vz^e$ and $\vz^c$ denote the respective values of $\rmZ^e$ and $\rmZ^c$. 

\emph{Text Generator.} For $p(\rmX_{ij}|  \rmZ^e, \rmZ^c)$, we considers a lightweight solution that only reconstructs a bag-of-words (BoW) representation from latent representations, which is significantly faster than a conventional sequence decoder.
\begin{equation}
\label{eq:pair_bow_prediction}
p(\vx^{\text{BoW}}|\vz^e, \vz^c) = \sigmoid(\boldsymbol{W}^{\text{dec}}[\vz^e, \vz^c]+ \boldsymbol{b}^{\text{dec}})
\end{equation}
where $\boldsymbol{\theta}_{\text{dec}}=[\boldsymbol{W}^{\text{dec}}; \boldsymbol{b}^{\text{dec}}]$ denotes the parameters of the decoder, $\sigma(\cdot)$ is the sigmoid function, and $\vx^{\text{BoW}}$ is the BoW representation of $\rmX_{ij}$.

\emph{Priors.} For both $p(\rmZ^e)$ and $p(\rmZ^c)$, we follow the common practice to use $\mathcal{N}(\mathbf{0}, \mathbf{I})$ as their priors.

\emph{Task-Specific Predictors.} For each predictor, we apply a linear layer to its inputs, followed by a softmax layer if it is a multi-class classification problem, otherwise a sigmoid layer for a binary classification problem.

\paragraph{Emotion Extraction Model.} We can apply any emotion extraction model to obtain clauses containing emotional expressions. In this work, we extend the emotion classification model in~\cite{xia2019emotion} by replacing its encoder with BERT encoder and its binary classification layer with a softmax layer.

\subsection{Model Training}
\label{sec:training}
\subsubsection{Source Domain Training}
\paragraph{\ourmodel Model.} Given a set of documents, each of which is annotated with a set $\gR^s_k=\{(y^r_{ij}, y^c_i, y^e_j)\}_{i,j}$ for positive examples, we obtain negative examples of relations by randomly sampling clause pairs that are not part of $\gR^s_k$. In particular, for each emotion clause in $\gR^s$, we pair it with a randomly picked non-cause clause in the document, resulting in the same number of negative samples. The training loss $\gL = \gL^{\text{ELBO}} + \lambda \Omega$, including the loss $\gL^{\text{ELBO}}$ derived from the ELBO and the variational posterior regularizer $\Omega$ adjusted by the hyperparameter $\lambda$.

Similar to prior works, the loss $\gL^{\text{ELBO}}$ includes the cross-entropy losses from the text decoder and the task-specific predictors, as well as two regularization terms from the two KL divergences, each of which takes the form of $\|\vz\|^2 - \log \vSigma$.

To motivate the regularizer $\Omega$, we start with Bhattacharyya distance, which measures the angle between two probability vectors $(\sqrt{p_a(z_0)}, ..., \sqrt{p_a(z_n)})$ and $(\sqrt{p_b(z_0)}, ..., \sqrt{p_b(z_n)})$ over $n$ data points. Unlike KL divergence, Bhattacharyya distance yields a positive value regardless the probability at a data point is zero or not, if the distance is not zero. For Gaussians, which are the cases for the variational posteriors, it has a closed form solution: 
\begin{eqnarray}
\label{eqn:variational_posteriors}
\fontsize{8pt}{8pt}\selectfont
\begin{aligned}
 \diver_{\text{bh}} = \frac{1}{8}(\vMu^e - \vMu^c)^{\text{T}}\mSigma^{-1} (\vMu^e - \vMu^c) + \frac{1}{2}\ln\big(\frac{\text{det}\mSigma}{\prod \sigma^e \prod\sigma^c}\big)
\end{aligned}
\end{eqnarray}
where $\mSigma = \frac{(\vSigma^e + \vSigma^c)^2}{2}\rmI$ and the determinant $\text{det}\mSigma = \frac{\prod((\sigma^e)^2 + (\sigma^c)^2)}{2}$. The left term is essentially an unnormalized multivariate Gaussian. The corresponding regularizer $\Omega^{\text{b}} = - \diver_{\text{bh}}$, which maximizes this distance, would drive the two Gaussians far away from each other.

 
The above regularizer only maximizes the distance between two types of latent representations from the same clause pair. Intuitively, it would be useful to also push $\vz^e_i$ of an instance $i$ away from the $\vz^c_j$ of the other instances. For efficiency, we only apply such regularizations between instances in a batch, which ends up a regularizer $\Omega^{\text{bb}}$ that maximizes Bhattacharyya distance between any pair of $(\vz^e_i, \vz^c_j)$ in a batch.
Following the same idea, we also exploit maximum mean discrepancy (MMD)~\cite{gretton2012kernel}, which is a kernel-based divergence measure not requiring absolute continuity, for maximizing divergences across instances batchwise. 

\begin{equation}
\label{eq:mmd_loss}
\begin{aligned}
\Omega^{\text{MMD}} = -&\Vert\phi(\vz^e)-\phi(\vz^c)\Vert^{2}_{\mathcal{H}},\\ &\vz^e\sim\rmZ^e, \vz^c\sim\rmZ^c
\end{aligned}
\end{equation}
where $\phi$ is a mapping function that projects both $\vz^e$ and $\vz^c$ into a reproducing kernel Hilbert space denoted by $\mathcal{H}$. In this work, we mainly adopt this regularizer in experiments due to its superior performance over the other two.
More in-depth discussion about the design of $Omega^{\text{MMD}}$ can be found in Appendix~\ref{appedix:sec1}.

\paragraph{Emotion Extraction Model.} Provided a set of clauses annotated with emotion categories or None, we train the emotion extraction model as a seven-way classification problem, following the maximum likelihood principle.

\subsubsection{Adaptation to Target Domains}
\label{sec:uda}
We transfer first the emotion extraction model to a target domain, followed by our model. The emotion extraction model is fine tuned by the self-training algorithm~\cite{chen2011co} on an unlabeled corpus in a target domain. The parameters of our model are fine tuned by using our method \ourmethod on the same corpora. Given an unlabeled corpus, both self-training algorithms start with applying the model to predict the most likely labels for each input text. The predictions are used to construct a training set to fine tune the model with the same loss $\gL$ as the source domain training in one epoch. Then the algorithms construct a new training set or update the training set with new examples by using the current model and repeats the process till the convergence criteria are met. Our algorithm \ourmethod differs from the current one in terms of the way to construct training datasets.

\emph{Relation Prediction.} Given a set of documents $\gD_u$ in a target domain, each of which contains at least one clause annotated with emotion pseudo-labels, we pair each emotion clause with the remaining clauses to create clause pairs for relation identification. When constructing a training set with pseudo-labels in each iteration, we select a pair with the highest probability in a document as a positive sample and randomly choose a clause pair from the remaining as a negative sample. Deep models with a high width tend to memorize training examples to reduce training errors~\cite{van2021memorization}, which could hurt the model performance by not improving its generalization capability. Thus, we construct a training set from scratch each time instead of updating the training set from the previous iteration. The training procedure terminates when a maximal number of iterations is reached.

\emph{Emotion Extraction.} For emotion extraction, we apply the self-training algorithm~\cite{chen2011co} to train the model in a target domain. It starts with an empty training set $\gD_t$ and a set of unlabeled documents $\gD_u$. In each iteration, if a document in $\gD_u$ contains at least one pseudo-labeled emotion clauses with their confidences above a pre-defined threshold, we add it to the training set $\gD_t$ for the next iteration. In each of such documents, we keep only the pseudo-labeled emotion clause with the highest probability, the remaining clauses are considered as non-emotion ones. 


\section{Experiments}
\subsection{Experimental Setup}
\label{exp_setup}


\textbf{Datasets.} Since there is no corpus for \targettask in the UDA setting, we divide \checpe into multiple domains.
Given the fact that the documents in \checpe are Chinese news articles sampled from the THUCNews dataset~\cite{li2007scalable}, we employ the topic classifier THUCTC~\cite{sun2016thuctc} trained on the THUCNews dataset to categorize \checpe into 14 subsets based on topics and choose the largest five as the final domains (e.g. home, society and finance, etc.). 
To further improve the purity of classification, based on THUCTC's classification results, we conduct manual inspection and labeling to complete the domain classification of \checpe.
Also, in the English language setting, we view \enecpe and Recognizing Emotion Cause in CONversations (RECCON)~\cite{DBLP:journals/cogcom/PoriaMHGBJHGRCG21} -- an English dataset specifically designed for identifying the causes of emotions within conversations, as the two source-target domains.
Table~\ref{table: corpus_stat} (in \ref{appendix:sec2}) summarizes the statistics of each corpus.

\paragraph{Metrics.} For each target domain in each corpus, we evaluate models for emotion extraction and relation identification respectively in terms of precision, recall and F1-score. A prediction is correct if there is a correct causal relation and the emotion category is correct.




\paragraph{Baselines.} To make a fair comparison, we adapt the three existing \targettask models RankCP, UTOS, UECA-Prompt (all employ $\text{BERT}$ as the backbone model) for emotion extraction (EE) and \targettask. 
In addition, since the universal prompt-based method for ECA tasks (UECA-Prompt)~\cite{DBLP:conf/coling/ZhengLZWW22} is designed to solve the different Emotion cause analysis (ECA) tasks in an unified framework, we thus only integrate three UDA approaches on the two ECPE models (RankCP~\cite{wei2020effective} and UTOS~\cite{cheng2021unified}) in the \targettask task to further demonstrate the effectiveness of our model.
The introduction of baseline method and implementation detail please refer to \ref{appendix:sec2}.

  
  
  


\subsection{Results and Analysis}
\label{sec:results}

\begin{table*}[hbt!]
\centering
\setlength{\tabcolsep}{12pt}
\Huge
\begin{adjustbox}{width=1\linewidth}
\begin{tabular}{c|cc|cc|cc|cc|cc}
\noalign{\hrule height 1.8pt}
\multirow{2}{*}{Model} & EE (\%) & ECPE (\%) & EE (\%) &  ECPE (\%) & EE (\%) &  ECPE (\%) & EE (\%) ECPE (\%) & EE (\%) & ECPE (\%)\\
                  & $F1$             & $F1$             & $F1$              & $F1$             & $F1$             & $F1$               & $F1$             & $F1$   
                  & $F1$ & $F1$\\
\hline                 
\textbf{(a) S: Society}       & \multicolumn{2}{c|}{Society $\rightarrow$ Home}                                              & \multicolumn{2}{c|}{Society $\rightarrow$ Finance}                                      & \multicolumn{2}{c|}{Society $\rightarrow$ Education        }                                       & \multicolumn{2}{c|}{Society $\rightarrow$ Entertainment}  & \multicolumn{2}{c}{Weighted Average}     \\
RankCP      &23.44
            &13.80

            &19.41
            &9.17

            &28.73
            &20.27

            &29.51
            &14.75
            
            &23.49 &13.65
            \\
RankCP+Ada-TSA    &19.77
                  &12.86			
                  
                  &16.61   
                  &7.42
                  
                  &22.41
                  &12.78

                  &25.21
                  &6.72
                  &19.65  &11.41
            \\  
RankCP+DANN      &\textbf{94.72}
                 &\textbf{66.67}			

                 &\textbf{88.96}
                 &52.58

                 &\textbf{87.21}
                 &54.42

                 &\textbf{83.05}
                 &40.00
                 &\textbf{92.03}  &60.93

            \\
RankCP+MEDM       &21.55
                  &13.39

                  &22.00
                  &10.46
                  
                  &26.78
                  &14.63
                  
                  &15.38
                  &6.84
                  
                  &22.04  &12.63

            \\
UTOS        &62.73
            &47.40

            &64.97
            &49.31

            &59.17
            &43.27

            &39.47
            &26.32

            &61.77 &46.39
            \\
UTOS+Ada-TSA      &19.77
                  &12.86
                  
                  &16.61
                  &7.42
                  
                  &22.41
                  &12.78
                  
                  &25.21
                  &6.72
                  &19.65  &11.41

            \\
UTOS+DANN        &71.10
                 &46.86
                  
                  &74.69
                  &48.36
                  
                  &68.89
                  &50.71
                  
                  &58.14
                  &34.88
                  &71.04  &47.16

            \\
UTOS+MEDM        &54.16
                 &20.31
                  
                 &26.15
                 &1.21

                 &46.01
                 &26.11
                  
                 &48.42
                 &15.49
                 &46.82  &16.70

            \\
            
UECA-Prompt &75.12
            &55.69

            &70.38
            &56.30
                  
            &79.17
            &54.97

            &72.22
            &51.56
            &74.48 &55.55
            \\
            
Ours        &78.85
            &64.60

            &83.88
            &\textbf{78.87}
                  
            &83.33
            &\textbf{76.88}
 
            &81.90
            &\textbf{82.24}
            &80.63 &\textbf{71.35}
            \\
\hline
\textbf{(b) S: Home}       & \multicolumn{2}{c|}{Home $\rightarrow$ Society}                                              & \multicolumn{2}{c|}{Home $\rightarrow$ Finance}                                      & \multicolumn{2}{c|}{Home $\rightarrow$ Education        }                                       & \multicolumn{2}{c|}{Home $\rightarrow$ Entertainment}               & \multicolumn{2}{c}{Weighted Average}     \\
RankCP            &\textbf{87.67}
                  &55.84

                  &\textbf{90.10}
                  &55.08

                  &\textbf{87.46}
                  &56.58

                  &\textbf{86.73}
                  &45.07

                  &\textbf{81.85} &51.31
                  \\
                  
RankCP+Ada-TSA    &17.75
                  &8.84
                  
                  &22.15
                  &8.75
                  
                  &20.06
                  &11.36
                  
                  &24.39
                  &6.50
                  &18.01  &8.40

            \\
RankCP+DANN      &32.87
                 &29.74
                  
                 &27.08
                 &15.91
                  
                 &35.79
                 &19.84
                  
                 &29.75
                 &14.88
                 &29.49  &22.73

            \\
RankCP+MEDM       &16.34
                  &7.20
                  
                  &7.54
                  &2.14
                  
                  &23.50
                  &9.09
                  
                  &25.42
                  &6.78
                  &14.55  &5.81

            \\
UTOS              &64.79
                  &51.16

                  &70.28
                  &48.82

                  &65.08
                  &49.41

                  &46.15
                  &30.77
                  &60.57 &45.89
                  \\
                  
UTOS+Ada-TSA      &17.75
                  &8.84
                  
                  &22.15
                  &8.75
                  
                  &20.06
                  &11.36
                  
                  &24.39
                  &6.50
                  &18.01  &8.40
            \\
            
UTOS+DANN         &73.41
                  &52.25
                  
                  &71.64
                  &50.63
                  
                  &68.38
                  &48.55

                  &57.47
                  &27.59
                  &66.45  &46.63

            \\
UTOS+MEDM         &44.67
                  &9.61

                  &24.11
                  &1.16

                  &45.64
                  &10.63
                  
                  &48.70
                  &13.04
                  &37.31  &7.37

            \\
UECA-Prompt       &80.57
                  &64.68

                  &80.07
                  &60.53

                  &78.24
                  &66.07

                  &75.23
                  &58.18

                  &69.10 &59.04
                  
                  \\
                  
Ours              &82.80
                  &\textbf{71.79}
                  
                  &83.22
                  &\textbf{79.54}
                  
                  &81.46
                  &\textbf{82.10}
                  
                  &81.13
                  &\textbf{82.57}
                  &76.72 &\textbf{70.09}
                  \\
\noalign{\hrule height 1.8pt}
\end{tabular}
\end{adjustbox}
\caption{Experimental results of our models and baselines utilizing F1 score (F1) as metrics on the ECPE and EE tasks on the Chinese dataset.
Emotion Extraction is denoted by EE while ECPE refers to Emotion-Cause Pair Extraction. S refers to source domain.}
\label{table: exp_results_brief}
\end{table*}

\begin{table*}[hbt!]
\centering
\fontsize{8}{10}\selectfont
\begin{adjustbox}{width=1\linewidth}
\begin{tabular}{c|cc|cc|cc}
\noalign{\hrule height 0.55pt}
\multirow{3}{*}{Model} & \multicolumn{2}{c|}{EN-ECPE $\rightarrow$ RECCON}                                              & \multicolumn{2}{c|}{RECCON $\rightarrow$ EN-ECPE}                            
& \multicolumn{2}{c}{Weighted Average}\\
                     & EE $F1$ (\%) & ECPE $F1$ (\%) & EE $F1$ (\%) & ECPE $F1$ (\%) & EE $F1$ (\%) & ECPE $F1$ (\%)\\
\hline                 
RankCP      &\textbf{39.86}
            &23.28
 
            &\textbf{52.96}
            &28.26
            
            &\textbf{47.87} &26.32
            \\
RankCP+Ada-TSA    &22.67
                  &12.13			
                  
                  &19.73   
                  &11.79
                  
                  &20.87  &11.92
            \\  
RankCP+DANN      &26.40
                 &14.87			

                 &32.17
                 &17.87
                 &29.93  &16.7

            \\
RankCP+MEDM       &21.79
                  &4.69

                  &30.15
                  &8.65
                  
                  &26.90  &7.11

            \\
UTOS        &33.96
            &27.83

            &24.13
            &18.48

            &27.95 &22.12
            \\
UTOS+Ada-TSA      &23.73
                  &11.21
                  
                  &19.13
                  &11.73

                  &20.92  &11.53

            \\
UTOS+DANN        &15.29
                 &3.36
                  
                  &13.91
                  &3.71
                  
                  &14.44  &3.57

            \\
UTOS+MEDM        &30.11
                 &1.55
                  
                 &18.09
                 &3.75

                 &22.76  &2.89

            \\
UECA-Prompt        &0.63
            &15.76

            &1.63
            &18.48
                  
            &1.24 &17.42
            \\
Ours        &29.57
            &\textbf{28.94}

            &21.58
            &\textbf{28.66}
                  
            &24.69 &\textbf{28.77}
            \\
\noalign{\hrule height 0.55pt}
\end{tabular}
\end{adjustbox}
\caption{Experimental results of our models and the baseline models on the English ECPE and EE tasks.}
\label{table: en_exp_results}
\end{table*}

\paragraph{Overall Comparisons.}
Table~\ref{table: exp_results_brief} (the ECPE and EE tasks on the Chinese dataset) and Table~\ref{table: en_exp_results} (the ECPE and EE tasks on the English dataset) report the results of our models and the baselines on the \targettask task, as well as the EE subtask.
For more detailed metrics on , i.e., precision (P) and recall (R) on Chinese dataset, please refer to Table~\ref{table: exp_results_large} in the Appendix section.
In Table~\ref{table: exp_results_brief} and \ref{table: en_exp_results}, the ``Weighted Average ECPE'' metric is the most important indicator of the model's cross-domain ECPE capability.

To dispel the doubt that our model outperforms the baselines only because they are developed in the supervised setting, we apply the SOTA UDA methods Ada-TS~\cite{zhang2021unsupervised}, DANN~\cite{ganin2016domain} and MEDM~\cite{wu2021entropy} to the two baselines RankCP and UTOS on the \task task.
MEDM is a minimal-entropy UDA approach that introduces diversity maximization to regulate entropy minimization for seeking a close-to-ideal domain adaptation.
Ada-TSA is a recently proposed adapter-based UDA approach in which the newly-added adapters can capture transferable features between source and target domains by using the domain-fusion scheme.
DANN is a widely adopted adversarial-based UDA approach that learns domain invariant representations through a domain discriminator.
It can be found that after applying the UDA framework, RankCP and UTOS significantly improved their performance and became comparable with the SOTA prompt-based model UECA-Prompt.  

However, though we employ UDA (for RankCP and UTOS) while leverage the powerful ability of the Large Language Model (LLM) (for UECA-Prompt) to enhance the baseline models, the baseline models still perform worse than our proposed model.
On \checpe, our model outperforms the RankCP+DANN by $10.42\%$ when treating society as the source domain, and UECA-Prompt by $11.05\%$ with home as the source domain in terms of weighted average F1.
On \enecpe, our model is better than the supervised learning model RankCP by  $2.45\%$.
Also, we can observe that our models get the best ECPE results in almost all of the domains except the $Society \rightarrow Home$ setting, indicating the generalization ability of the proposed approach. 
It is worth mentioning that our model performs the best even it does not always achieve the best performance on the EE subtask.

Note that there is a significant performance gap between the Chinese and English benchmarks. The cause of this gap mainly due to the distribution bias problem where the five domains used for testing in the Chinese benchmark are extracted from the same corpus, i.e., CH-ECPE, however the two domains under the English setting derive from the two different datasets RECCON and EN-ECPE. Therefore, compared with the Chinese domains, the two English domains share less knowledge between each other, making the model hard to transfer from one domain to another. 

To sum up, for the Weighted Average ECPE metric, which is the most important indicator of the model's cross-domain ECPE capability, RankCP and UTOS, even when using domain adaptation techniques (MEDM, ada-TSA, and DANN), performed lower than our model on both Chinese and English datasets. 
This demonstrates that our method can effectively enable the model to learn cross-domain ECPE capabilities, and prove the strengths of our model in terms of identifying new causal relations between events and emotions in new domains.

\begin{table*}[hbt!]
\centering
\fontsize{8}{10}\selectfont
\begin{adjustbox}{width=1\linewidth}
\begin{tabular}{c|ccc|ccc|ccc|ccc}
\noalign{\hrule height 1pt}
\multirow{3}{*}{Model} & \multicolumn{3}{c|}{Society $\rightarrow$ Entertainment}                                              & \multicolumn{3}{c|}{Society $\rightarrow$ Home}                                              & \multicolumn{3}{c|}{Society $\rightarrow$ Education}                                               & \multicolumn{3}{c}{Society $\rightarrow$ Finance}                                               \\
                  & \multicolumn{3}{c|}{ECPE (\%)} &  \multicolumn{3}{c|}{ECPE (\%)} &  \multicolumn{3}{c|}{ECPE (\%)} &  \multicolumn{3}{c}{ECPE (\%)}  \\
                 & $P$ & $R$ & $F1$               & $P$ & $R$ & $F1$             & $P$ & $R$ & $F1$              & $P$ & $R$ & $F1$                                                     \\
\hline
Original  &84.62 &80.00 &82.24

            &58.59 &71.98 &64.60

            &74.30 &79.64 &76.88

            &75.96 &82.01 &78.87
            \\
\hline
w/o MMD     &69.63 &74.02 &71.76

            &49.77 &48.70 &49.23

            &65.54 &69.78 &67.60

            &68.65 &58.63 &63.30
            \\
w/o HSIC        &59.87 &73.23 &65.88

            &40.51 &51.76 &45.66

            &61.73 &73.38 &67.05

            &64.23 &61.57 &62.88
            \\
w/o VI          &63.51 &74.02 &68.36

            &45.97 &52.52 &49.09

            &60.24 &71.94 &65.57

            &69.12 &60.59 &64.61

            \\
w/o $\Omega^b$    &61.66 &61.42 &61.54

            &39.50 &55.57 &46.58

            &62.91 &76.26 &68.94

            &60.31 &67.45 &63.71

            \\
w/o $\Omega^{bb}$    &76.52 &79.53 &77.99

            &54.80 &52.52 &53.64

            &66.55 &71.49 &68.95

            &83.10 &69.41 &75.71
            \\
w/o $\Omega^{\text{MMD}}$         
            &78.12 &78.74 &78.43

            &64.30 &57.86 &60.95

            &69.14 &80.58 &74.42

            &86.39 &68.43 &76.49
            \\
\hline
w/o Adapter &86.67 &75.00 &80.44

            &59.05 &71.16 &64.54

            &75.88 &74.44 &75.15

            &75.74 &79.93 &77.78
            \\
w/o Self-training &45.24 &34.55 &39.18
           
            &18.63 &66.00 &29.06
                  
            &25.62 &61.68 &36.20
 
            &27.19 &51.56 &35.60
            \\ 
\hline
with Gold Emotions &89.83 &96.36 &92.98
           
            &78.32 &89.80 &83.67
                  
            &90.48 &91.02 &90.75
 
            &74.16 &91.35 &81.86
            \\ 
\noalign{\hrule height 1pt}
\end{tabular}
\end{adjustbox}
\caption{Experimental results of our models with different settings for the \targettask task on CH-ECPE.}
\label{table: exp_results_regularization}
\end{table*}

\paragraph{Ablation Study.} To analyze the influence that different module might exert on the proposed approach, we conduct the ablation study. The second row (named `Original') in Table \ref{table: exp_results_regularization} refers to the result that our model could get when it is equipped with all the techniques presented in this work.

To study the effect of the regularizer $\Omega$ (see Sec. \ref{sec:uda}) for disentangled representation learning, we remove the $\Omega^{\text{MMD}}$ during model training, as well as compare it with the other types of regularizers, including two independence measures Hilbert–Schmidt independence criterion~\cite[(HSIC]{gretton2005measuring} and Variation of Information~\cite[(VI]{cheng2020improving}. From Table \ref{table: exp_results_regularization} we can see that there is at least a 2.38\% drop in terms of F1 on \checpe when the regularizer $\Omega^{\text{MMD}}$ is removed. Adding HSIC does more harm than gain, and VI brings almost no benefits to the model. It is also not useful to only apply the regularizer $\Omega^b$, which maximizes Bhattacharyya distance between the variational posteriors $q(\rmZ^e| \rmX_{ij})$ and $q(\rmZ^c| \rmX_{uv})$ from the same clause pair. However, the regularizer works when we maximize Bhattacharyya distance between two variational posteriors from all possible instance pairs in a batch. Similarly, the MMD-based regularizer $\Omega^{\text{MMD}}$ works also because it maximizes the MMD distance across instances. 

Also, we remove Emotion and Event adapters and use the unified pair representation as the input for both the emotion and event encoders. By doing this we lost performance for all domains, as the Table \ref{table: exp_results_regularization} shows. It is proved that using the different vectors to represent the emotion / event variables is a better solution.
In addition, we also conduct experiments on investigating the efficacy of self-training and regularizer, detailed in \ref{appendix:sec3}. 

\section{Related Work}
\label{related_work}

\paragraph{Emotion-Cause Pair Extraction.}
ECPE is a new task that aims to extract all potential emotions and corresponding causes in a unannotated document.
The pioneer~\cite{xia2019emotion} proposes a two-step approach that first extracts emotion and cause clauses separately.
~\citet{wei2020effective} propose a joint neural approach that applies graph attention to model the interrelations between clauses and rank ECPE.
~\citet{DBLP:conf/coling/ZhengLZWW22} first introduce prompt learning method into the ECPE task by decomposing the ECPE task into multiple sub-tasks and design prompts for each the sub-task.

Our model is different from existing works in two main aspects.
Firstly, we tackle ECPE in the UDA setting, which is more difficult and practical as it allows distribution discrepancies between different domains.
Secondly, we solve UDA-ECPE from a causal perspective and design a causal disentanglement mechanism to approximate emotion and cause random variables, enabling causal discovery to identify causal relations between them and consequently retrieve positive pairs.

\paragraph{Unsupervised Domain Adaptation.}
Domain adaptation addresses domain shift, allowing a pre-trained model to generalize from a source to a target domain. It falls into two types: supervised and unsupervised(examples of both types can be found in \ref{appendix:sec4}).

Our work focuses on unsupervised domain adaptation (UDA), specifically extracting cross-domain emotion-cause pairs from labeled source domains to unlabeled target domains. Unlike prior studies~\cite{miller2019simplified,du2020adversarial,zou2021unsupervised,karouzos2021udalm,zhang2021unsupervised} on binary sentiment classification, we tackle non-binary variables (emotion and cause) that are causally linked. This is the first known attempt to discover causal relations in UDA.



\paragraph{Disentangled Representation Learning.} 
The aim of disentangled representation learning (DRL) is to learn factorized representations that reveal the semantically meaningful factors hidden in the observed data~\cite{bengio2013representation,higgins2018towards}.
Mainstream DRL approaches in NLP~\cite{john2019disentangled,cheng2020improving,vishnubhotla2021evaluation} learn such representations by adopting variational autoencoders ~\cite[VAE]{kingma2013auto}, which achieve disentanglement via the Kullback-Leibler~\cite[KL]{kullback1951information} divergence minimization between the posterior of the latent factors and a standard multivariate normal prior. 
Additionally, a deep VAE model with innovative disentanglement priors, named \normalsize{VAEDPRIOR}, is proposed for task-specific natural language generation in zero-shot or few-shot scenarios~\cite{li2022variational}.


\paragraph{Can LLMs well solve the causal discovery tasks?} 
Despite their advanced linguistic capabilities and technological breakthroughs, LLMs struggle with causal inference in situations where the variable names and textual expressions in queries differ from those in their training data~\cite{DBLP:journals/tmlr/ZecevicWDK23,DBLP:conf/iclr/Jin0LPSMDS24}. 
The ability of LLMs to conduct causal discovery remains a subject of debate. 
For instance, researchers demonstrate that in specialized domains such as medicine and climate science, LLMs can accurately determine pairwise causal relationships, achieving accuracies as high as 97\%, albeit this success often depends on carefully tailored prompts~\cite{kiciman2023causal}. 
However, in other real-world domains, smaller, specialized models consistently outperform GPT-3 and GPT-4 in event causality identification (ECI) tasks—those that pinpoint cause/effect spans in text descriptions~\cite{DBLP:conf/naacl/GaoCH19}. 
These smaller and specialized models also greatly surpass LLMs in binary pairwise causality inference~\cite{DBLP:conf/emnlp/RomanouMPLAB23}. 

Given their training on vast quantities of natural language texts, LLMs are proficient at recognizing causal event pairs but falter with non-causal relationships, which raises concerns about their tendency to memorize rather than generalize event knowledge~\cite{DBLP:conf/emnlp/RomanouMPLAB23,DBLP:conf/emnlp/JacoviCGG23,DBLP:journals/corr/abs-2407-19638}. 
Specifically, ChatGPT has a serious hallucination on causal reasoning, making it an inadequate causal reasoner~\cite{DBLP:conf/emnlp/GaoD0023}. 

In summary, empirical research indicates that LLMs still exhibit deficiencies in causal discovery tasks, and there is ongoing debate about their effectiveness in handling causal discovery and reasoning. Consequently, we did not use LLMs as baseline models for comparison in this work. However, future research will include rigorous experiments and comparisons to assess LLMs' performance in causal discovery tasks.

\section{Conclusion}
We propose a novel causal discovery inspired VAE model and a customized self-training algorithm for the \task task. Herein, we propose to disentangle the latent representations of emotions from those of events by a novel variational posterior regularization technique that does not enforce independence between the corresponding latent random variables. This work also sheds the light on the connections between the task of causal relation identification in the NLP community and the causal discovery theory, paves the way for theoretically grounded approaches to comprehensively analyzing causal structures in texts.
\section*{Limitations}
A potential limitation of this work is that, due to resource and time constraints, we only used the ECPE classification model based on Bert, which matches our model's architecture, as the baseline model. We did not compare it with the latest large language models (LLMs). Recent studies indicate that LLMs are not particularly effective at solving causal discovery tasks. Therefore, in the future, we plan to include the following LLM-based baseline models: zero-shot learning-based LLM (encapsulating the ECPE task in a task instruction prompt to obtain answers from the LLM), few-shot learning-based LLM (selecting a few ECPE examples as in-context learning demonstrations), and SFT-based LLM (fine-tuning the LLM using the ECPE dataset as task instruction). In future work, we will compare the method proposed in this paper with LLM-based methods to empirically explore whether LLM models can be effectively applied to causal discovery tasks.
\section*{Ethics Statement}
This research involves the development of a model for emotion-cause pair extraction in an unsupervised domain adaptation setting, which carries certain ethical considerations. 

First, our model relies on datasets that may contain biases inherent in the language, emotional expressions, or cultural differences, which could inadvertently lead to biased outputs.
We have taken precautions to manually mitigate these biases by employing human annotators. 
However, we acknowledge that some biases may still remain. 

Additionally, the use of causal discovery methods must be carefully considered, as incorrect identification of causal relationships could lead to misguided conclusions in real-world applications. 

Also, we have released the source code and dataset, and we urge future researchers and practitioners to use the code responsibly, respecting data privacy and avoiding any misuse that could lead to negative social impacts.

Furthermore, we employed human annotators to improve the quality of our training and test datasets. We ensured that all annotators were informed about the purpose of the data annotation and that their work was conducted under fair labor practices, including appropriate compensation and voluntary participation.
\section*{Acknowledgments}
This work is partly supported by the ARC Future Fellowship FT190100039.
This material is based on research sponsored by DARPA under agreement number HR001122C0029 (CCU Program). 
The U.S. Government is authorized to reproduce and distribute reprints for Governmental purposes notwithstanding any copyright notation thereon.

\clearpage
\balance
\bibliography{main}
\clearpage
\appendix
\section{Appendix}
\label{sec:appendix}

\subsection{Visualization of sentence embeddings for English \task corpora}
\label{appedix:sec1}
\begin{figure*}[hbt!]
     \centering
     \begin{subfigure}[b]{0.4\textwidth}
         \centering
         \includegraphics[width=\textwidth]{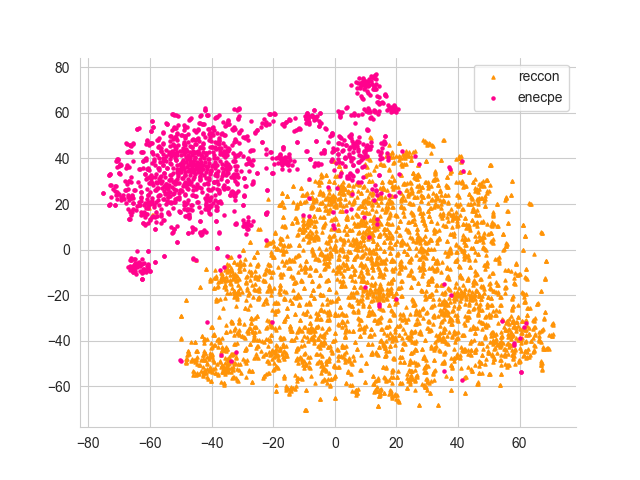}
         \caption{English emotion cause clauses}
         \label{fig: en_emo}
     \end{subfigure}
     \hfill
     \begin{subfigure}[b]{0.4\textwidth}
         \centering
         \includegraphics[width=\textwidth]{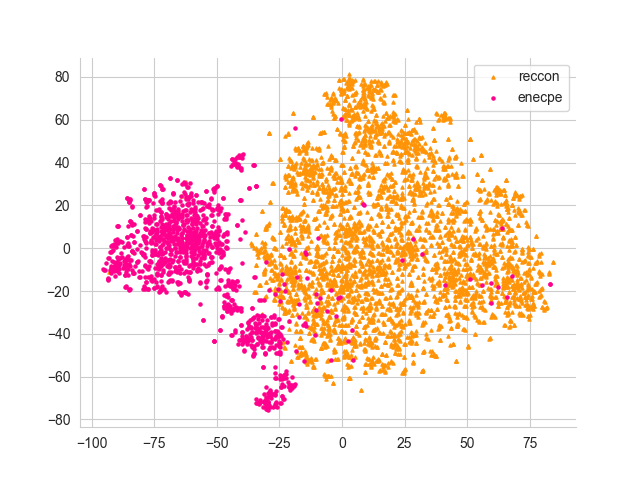}
         \caption{English emotion clauses}
         \label{fig: en_cause}
     \end{subfigure}
     \hfill

     \caption{The t-SNE visualizations of the clause embeddings from the English \task corpora }
     \label{fig:EN-ECPE_challenges}
\end{figure*}

As shown in Fig.\ref{fig: en_emo} and Fig.\ref{fig: en_cause}, regardless if a clause mentions an emotion or an emotion cause, there is a very clear boundary between the two domains. Their domain differences are largely caused by the differences between the two datasets.

\subsection{The essence of machine learning and mathematical considerations in designing the regularization term $\Omega$}
\label{appedix:omega}
In the original design of the VAE's ELBO, we need to minimize $\KL(q(\rmZ^e | \rmX_{ij} \| p(\rmZ^e)) $ and $\KL(q(\rmZ^c | \rmX_{ij} \| p(\rmZ^c))$.
This is to minimize the difference between the posterior distribution of the latent variables output by the encoder and the prior distribution, ensuring that the latent variable distribution has reasonable properties while learning the data generation process.

In our design, we introduced a regularizer $\Omega$, to reduce the overlap between $q(\rmZ^e | \rmX_{ij})$ and $q(\rmZ^c | \rmX_{ij})$. 
This means increasing the divergence between the posterior distributions of the latent variables output by the emotion and event encoders to minimize the overlap.

The above two points are not contradictory. 
By combining the ELBO and the regularizer loss, we can train our model to make the posterior distributions of the encoder's latent variables, $q(\rmZ^e | \rmX_{ij})$ and $q(\rmZ^c | \rmX_{ij})$, respectively closer to the prior distributions $p(\rmZ^e)$ and $p(\rmZ^c)$. 

Under this constraint, we can increase the ``distance'' between the posterior distributions of the two encoders' latent variables, i.e., $q(\rmZ^e | \rmX_{ij})$ and $q(\rmZ^c | \rmX_{ij})$. 
Since the prior distributions $p(\rmZ^e)$ and $p(\rmZ^c)$ are two predefined multivariate Gaussian distributions that are independent of each other, we do not need to constrain the distance between these two prior distributions. 

Therefore, we can consider $p(\rmZ^e)$ and $p(\rmZ^c)$ as fixed in the vector space, and minimizing $\KL(q(\rmZ^e | \rmX_{ij} \| p(\rmZ^e)) $ and $\KL(q(\rmZ^c | \rmX_{ij} \| p(\rmZ^c))$ is to make the posterior distributions in the vector space as close as possible to $p(\rmZ^e)$ and $p(\rmZ^c)$. Maximizing the distance between $q(\rmZ^e | \rmX_{ij})$ and $q(\rmZ^c | \rmX_{ij})$ is to increase the distance between these two posterior distributions while they are close to $p(\rmZ^e)$ and $p(\rmZ^c)$.

Based on the above discussion, the design of the ELBO and the regularizer $\Omega$ does not conflict.

However, of course, we must recognize that the posterior distributions of $p(\rmZ^e)$ and $p(\rmZ^c)$ are both close to a multivariate normal distribution following $N(0, I)$, which may lead to conflicts due to their overly similar distributions, resulting in reduced distinguishability. 
We can use different model structures to generate $p(\rmZ^e)$ and $p(\rmZ^c)$ that are as different as possible in distribution for the same input $\rmX_{ij}$, thereby reducing the risk of conflict. 
We plan to conduct such modeling in future work to further empirically study the issue of posterior distribution conflict.

\subsection{Baseline Model and Implementation Detail}
\label{appendix:sec2}
\begin{table}[hbt!]
\centering
\small
\begin{tabular}{ccc} 
\noalign{\hrule height 1.8pt}
Language                 & Domain        & \#Docs  \\ 
\hline
\multirow{5}{*}{Chinese} & Home       &  746 \\
                         & Society & 659     \\
                         & Finance          & 263     \\
                         & Education     & 153     \\
                         & Entertainment       & 52      \\ 
\hline
\multirow{2}{*}{English} & \enecpe   & 1226 \\
                             & RECCON       & 780 \\
\noalign{\hrule height 1.8pt}
\end{tabular}
\caption{The statistics of the \task corpora.}
\label{table: corpus_stat}
\end{table}

\textbf{RankCP} performs the emotion-cause pair extraction using the graph attention network, which models the inter-clause information and extracts the valid emotion-cause pairs from a ranking perspective.
  
\textbf{UTOS} adopts the unified sequence labeling approach to extract emotion-cause pairs in a way that the position of emotion and cause clauses as well as how they pair can be predicted via one pass of sequence labeling.
  

\textbf{UECA-Prompt} designs sub-propmts for the emotion extraction, cause extraction, and emotion-cause pair extraction sub-tasks, then synthesize the sub-prompts to solve the ECA task.

\textbf{Implementation Details.} We adopt $\text{BERT}_{ZH}$\footnote{\url{https://huggingface.co/hfl/chinese-roberta-wwm-ext}} and $\text{BERT}_{EN}$\footnote{\url{https://huggingface.co/roberta-base}} as the clause pair encoders for Chinese and English, respectively.
The size of hidden bidirectional LSTM in emotion extraction model is set to 100.
The outputted dimensions of emotion classifier and event predictor in \ourmodel are set to 24.
The confidence threshold for the self-training of emotion extraction model is set to 0.7.
The number of iterations for the self-training of event-emotion relation model is set to 50.

We train the emotion extraction model and the \ourmodel by using Adam optimizer, where the learning rates and the mini-batch sizes are 2e-5 and 4 and 1e-5 and 64, respectively.
As for regularization, we apply dropout to both of them with the dropout rate 0.5.

\subsection{Ablation Study in Self Training}
\label{appendix:sec3}

\begin{figure*}[hbt!]
\centering
\includegraphics[width=1.0\textwidth]{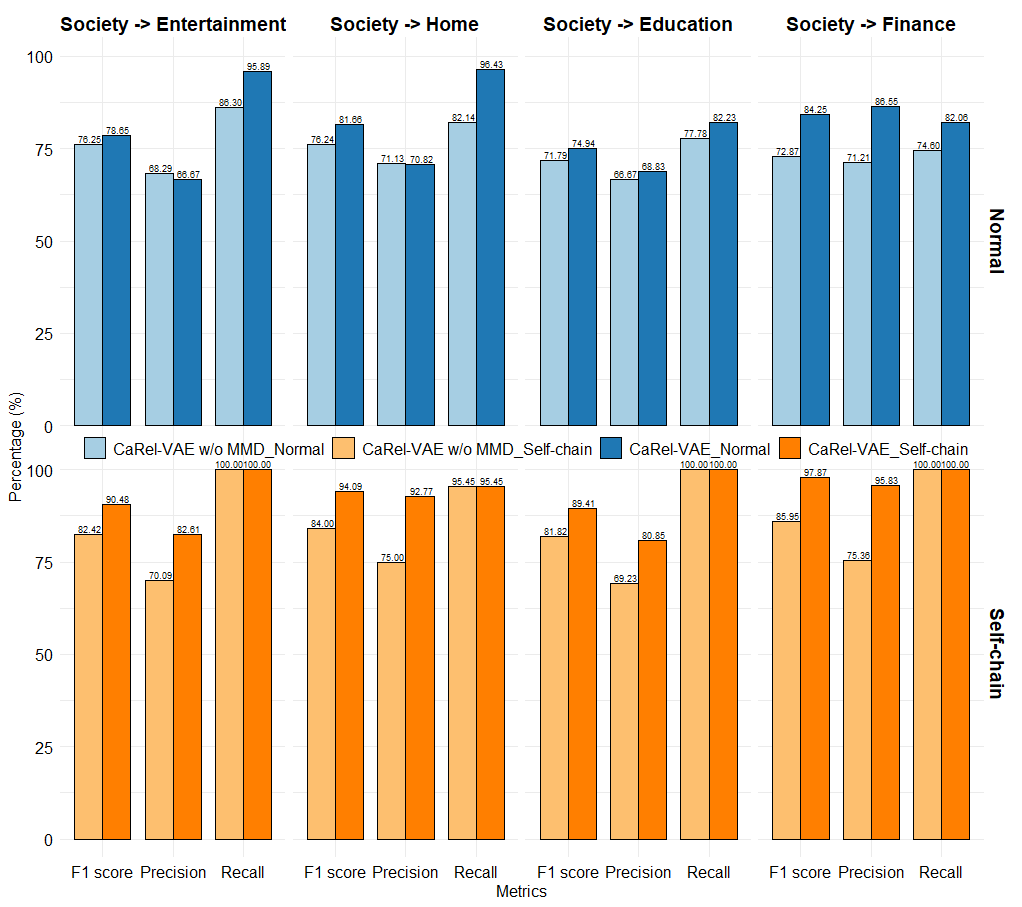}
\caption{Experimental results of \ourmodel w/o MMD and \ourmodel for normal and self-chain cases. The normal case refers to an emotion-cause pair composed of two different clauses, while for the self-chain case a pair are mentioned in the same clause.}
\label{fig:normal_self_results}
\end{figure*}

We train the model using the source domain's ground-truth labels, and then directly apply this supervised-learning model to the target domain without any self-training. In the `w/o Self-training' row of the Table \ref{table: exp_results_regularization}, we can see the model experiences a major performance drop, indicating the usefulness of the self-training.

Furthermore, it is also interesting to explore the extent to which the predicted emotion labels, aka EE's results, will influence the downstream ECPE's performance. We therefore utilize the ground-truth emotion labels instead of the ones that are predicted by the emotion extraction model as the input of the ECPE task. In the last row of the Table \ref{table: exp_results_regularization}, the minimum improvement observed is 2.99\% in terms of F1 among all domains, showing that the quality of the emotion prediction does have a certain impact on the ECPE task. However, our model can still achieve the best results even we only use an emotion extraction model with a moderate performance to predict the emotions, whose task is not the focus of this work.


\paragraph{Regularizer.} To further understand how $\Omega^{\text{MMD}}$ contributes to the \task task, we examine the performance of our original model and its variant for two different types of emotion-cause pairs including normal and self-chain, the results are shown in Figure \ref{fig:normal_self_results}.
Observe that the performance improvement is mainly attributed to the significant increment of precision in self-chain cases.
This suggests that disentangled representation learning helps approximate emotion and cause random variables from emotion-cause pairs, and ultimately aids in the causal discovery process.



\begin{figure*}[hbt!]
\centering
\includegraphics[width=0.9\textwidth]{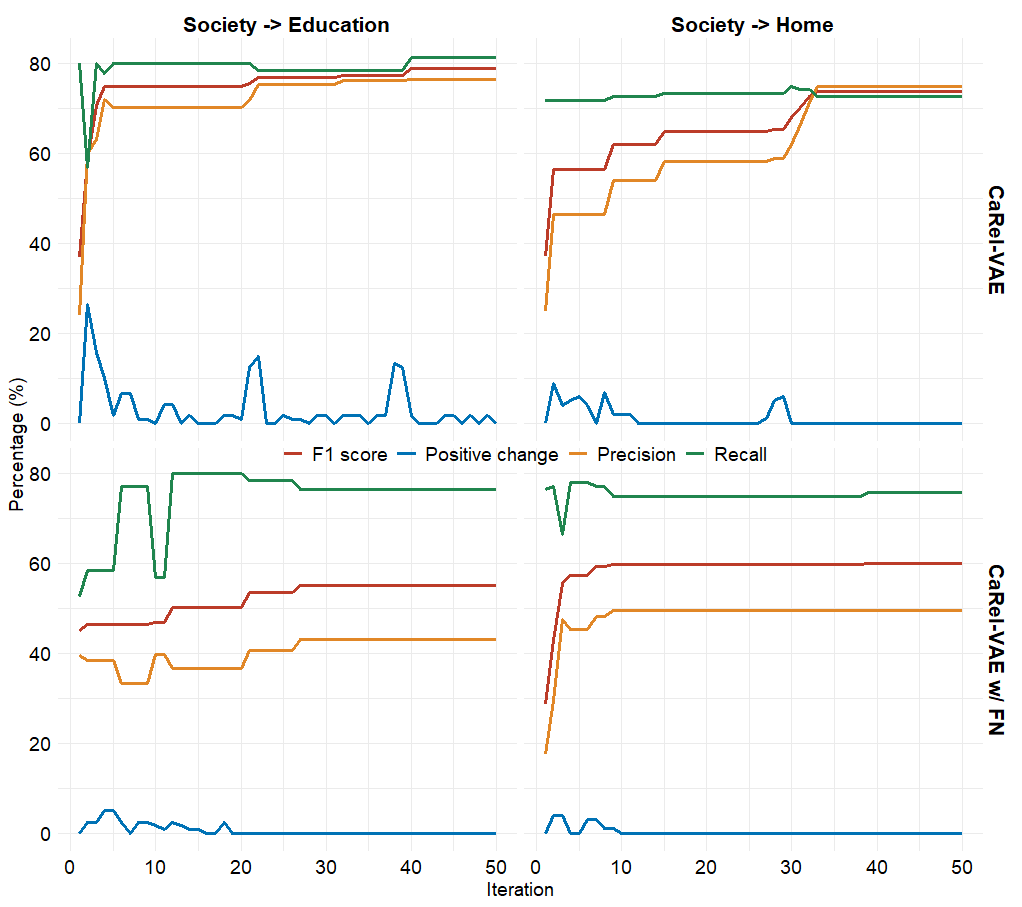}
\caption{Experimental results of our variant models that fixes negative samples during the self-training (denoted as "\ourmodel w/ FN") and our original model \ourmodel.}
\label{fig:self_train_random_editting_analysis}
\end{figure*}




\paragraph{Improved Self-training.} For \ourmethod, we examine the usefulness of always constructing a new training set in each iteration during self-training. As a comparison, we only update the training set from the previous iteration by adding new documents. In this way, negative examples in the training set remain the same once their documents are added to the training set. Fig.~\ref{fig:self_train_random_editting_analysis} reports the proportion of changed positive examples and the proportion of changed examples in each iteration, as well as changes of precision/recall/F1 over time. We can see that changing negative examples in each iteration indeed prevents the model from memorizing the training examples so that it improves the generalization capability of our model.
\subsection{Additional Content for related work}
\label{appendix:sec4}
Depending on the situation of target domain data, Domain adaptation can be categorized into two broad
classes: supervised domain adaptation and unsupervised domain adaptation.
The former can achieve promising results given the small amount of target domain labeled data~\cite{daume2007frustratingly,plank2011domain}.
Conversely, the unsupervised domain adaptation (UDA) does not require any data in the target domain to be labeled and thus is more attractive and challenging~\cite{glorot2011domain,ramponi2020neural}.
Our work falls under the UDA research area.
Specifically, cross-domain emotion-cause pair extraction from one source domain with labels to various unlabeled target domains.
Unlike most previous works~\cite{miller2019simplified,du2020adversarial,zou2021unsupervised,karouzos2021udalm,zhang2021unsupervised} on cross-domain sentiment classification that solely work with a binary categorical variable (i.e., positive or negative sentiment), we simultaneously focus on two non-binary ones (i.e., emotion and cause) that are causally dependent.
To the best of our knowledge,  this is the first attempt at discovering causal relations in the context of UDA.

\clearpage

\begin{table*}[hbt!]
\centering
\setlength{\tabcolsep}{12pt}
\Huge
\begin{adjustbox}{width=1\linewidth}
\begin{tabular}{c|cccccc|cccccc}
\noalign{\hrule height 1.8pt}
\multirow{2}{*}{Model} & \multicolumn{3}{c}{EE (\%)} & \multicolumn{3}{c|}{ECPE (\%)} & \multicolumn{3}{c}{EE (\%)} &  \multicolumn{3}{c}{ECPE (\%)} \\
                  & $P$ & $R$ & $F1$             & $P$ & $R$ & $F1$             & $P$ & $R$ & $F1$              & $P$ & $R$ & $F1$ \\
\hline                 
\textbf{(a) S: Society}       & \multicolumn{6}{c|}{Society $\rightarrow$ Home}                                              & \multicolumn{6}{c}{Society $\rightarrow$ Finance}  \\
RankCP      &21.90 &25.22 &23.44
            &13.14 &14.54 &13.80

            &18.04 &21.00 &19.41
            &8.56 &9.86 &9.17

            \\
RankCP+Ada-TSA    &18.55 &21.16 &19.77
                  &12.30 &13.48 &12.86			
                  
                  &15.86 &17.44 &16.61   
                  &7.12 &7.75 &7.42
                  
                  \\  
RankCP+DANN      &\textbf{91.51} &\textbf{98.15} &\textbf{94.72}
                 &51.69 &\textbf{93.85} &\textbf{66.67}			

                 &85.06 &\textbf{93.24} &\textbf{88.96}
                 &40.38 &75.35 &52.58

                 \\
RankCP+MEDM       &20.17 &23.12 &21.55
                  &12.77 &14.07 &13.39

                  &20.43 &23.84 &22.00
                  &9.76 &11.27 &10.46
            \\
UTOS        &\textbf{91.51} &47.72 &62.73
            &\textbf{70.99} &35.58 &47.40

            &\textbf{93.33} &49.82 &64.97
            &71.33 &37.68 &49.31
            \\
UTOS+Ada-TSA      &18.55 &21.16 &19.77
                  &12.30 &13.48 &12.86
                  
                  &15.86 &17.44 &16.61
                  &7.12 &7.75 &7.42
                  
            \\
UTOS+DANN        &84.96 &61.13 &71.10
                 &56.41 &40.07 &46.86
                  
                  &89.55 &64.06 &74.69
                  &57.84 &41.55 &48.36
            \\
UTOS+MEDM        &52.80 &55.60 &54.16
                 &14.63 &33.32 &20.31
                  
                 &15.31 &89.68 &26.15
                 &0.64 &13.03 &1.21
            \\
UECA-Prompt        &75.59 &74.66 &75.12
            &50.92 &61.43 &55.69

            &71.01 &69.75 &70.38
            &51.13 &62.63 &56.30
            \\
Ours        &81.77 &76.14 &78.85
            &58.59 &71.98 &64.60

            &86.42 &81.49 &83.88
            &\textbf{75.96} &\textbf{82.01} &\textbf{78.87}
            \\
            
\hline

\textbf{(a) S: Society}      & \multicolumn{6}{c|}{Society $\rightarrow$ Education}                                       & \multicolumn{6}{c}{Society $\rightarrow$ Entertainment}\\
RankCP      &26.13 &31.90 &28.73
            &18.59 &22.29 &20.27

            &26.87 &32.73 &29.51
            &13.43 &16.36 &14.75
            \\
RankCP+Ada-TSA    &20.62 &24.54 &22.41
                  &11.86 &13.86 &12.78

                  &23.44 &27.27 &25.21
                  &6.25 &7.27 &6.72
            \\  
RankCP+DANN      &82.87 &\textbf{92.02} &\textbf{87.21}
                 &43.01 &74.10 &54.42

                 &77.78 &\textbf{89.09} &\textbf{83.05}
                 &30.48 &58.18 &40.00
            \\
RankCP+MEDM       &24.14 &30.06 &26.78
                  &13.30 &16.27 &14.63
                  
                  &14.52 &16.36 &15.38
                  &6.45 &7.27 &6.84
            \\
UTOS        &\textbf{92.21} &43.56 &59.17
            &67.09 &31.93 &43.27

            &71.43 &27.27 &39.47
            &47.62 &18.18 &26.32
            \\
UTOS+Ada-TSA      &20.62 &24.54 &22.41
                  &11.86 &13.86 &12.78
                  
                  &23.44 &27.27 &25.21
                  &6.25 &7.27 &6.72

            \\
UTOS+DANN        &86.92 &57.06 &68.89
                  &62.28 &42.77 &50.71
                  
                  &80.65 &45.45 &58.14
                  &48.39 &27.27 &34.88

            \\
UTOS+MEDM        &53.00 &62.50 &46.01
                 &24.23 &28.31 &26.11
                  
                 &57.50 &41.82 &48.42
                 &12.64 &20.00 &15.49

            \\
UECA-Prompt &75.84 &82.82 &79.17
            &48.84 &62.87 &54.97

            &73.58 &70.91 &72.22
            &45.21 &60.00 &51.56
            \\
Ours        &83.85 &82.82 &83.33
            &\textbf{74.30} &\textbf{79.64} &\textbf{76.88}
 
            &\textbf{86.00} &78.18 &81.90
            &\textbf{84.62} &\textbf{80.00} &\textbf{82.24}
            \\
\hline
\hline
\textbf{(b) S: Home}       & \multicolumn{6}{c|}{Home $\rightarrow$ Society}                                              & \multicolumn{6}{c}{Home $\rightarrow$ Finance}     \\
RankCP            &83.88 &\textbf{91.82} &\textbf{87.67}
                  &44.33 &\textbf{75.42} &55.84

                  &86.56 &\textbf{93.95} &\textbf{90.10}
                  &43.41 &75.35 &55.08
                  \\
RankCP+Ada-TSA    &16.38 &19.37 &17.75
                  &8.25 &9.51 &8.84
                  
                  &20.42 &24.20 &22.15
                  &8.11 &9.51 &8.75

            \\
RankCP+DANN      &29.29 &37.45 &32.87
                 &26.79 &33.43 &29.74
                  
                 &25.00 &29.54 &27.08
                 &14.76 &17.25 &15.91

            \\
RankCP+MEDM        &15.43 &17.36 &16.34
                  &6.89 &7.55 &7.20
                  
                  &7.61 &7.47 &7.54
                  &2.17 &2.11 &2.14
            \\
UTOS              &88.56 &51.08 &64.79
                  &\textbf{70.69} &40.08 &51.16

                  &\textbf{90.00} &57.65 &70.28
                  &62.30 &40.14 &48.82
                  \\
UTOS+Ada-TSA      &16.38 &19.37 &17.75
                  &8.25 &9.51 &8.84
                  
                  &20.42 &24.20 &22.15
                  &8.11 &9.51 &8.75
            \\
UTOS+DANN         &\textbf{87.98} &62.98 &73.41
                  &63.04 &44.62 &52.25
                  
                  &89.36 &59.79 &71.64
                  &63.16 &42.25 &50.63
            \\
UTOS+MEDM         &33.96 &65.28 &44.67
                  &5.52 &37.20 &9.61

                  &13.85 &92.88 &24.11
                  &0.61 &14.44 &1.16
            \\
UECA-Prompt              &76.52 &85.08 &80.57
                  &66.33 &63.11 &64.68

                  &78.04 &82.21 &80.07
                  &61.96 &59.17 &60.53
                  \\
Ours              &86.07 &79.77 &82.80
                  &68.78 &75.07 &\textbf{71.79}
                  
                  &81.79 &84.70 &83.22
                  &\textbf{76.03} &\textbf{83.39} &\textbf{79.54}
                  \\
\hline
\textbf{(b) S: Home}    &   \multicolumn{6}{c|}{Home $\rightarrow$ Education}                                       & \multicolumn{6}{c}{Home $\rightarrow$ Entertainment}  \\
RankCP            &83.33 &\textbf{92.02} &\textbf{87.46}
                  &44.48 &77.71 &56.58

                  &\textbf{84.48} &\textbf{89.09} &\textbf{86.73}
                  &36.78 &58.18 &45.07

                  \\
RankCP+Ada-TSA    &18.82 &21.47 &20.06
                  &10.75 &12.05 &11.36
                  
                  &22.06 &27.27 &24.39
                  &5.88 &7.27 &6.50

            \\
RankCP+DANN      &31.34 &41.72 &35.79
                 &17.51 &22.89 &19.84
                  
                 &27.27 &32.73 &29.75
                 &13.64 &16.36 &14.88

            \\
RankCP+MEDM       &22.04 &25.15 &23.50
                  &8.60 &9.64 &9.09
                  
                  &23.81 &27.27 &25.42
                  &6.35 &7.27 &6.78

            \\
UTOS              &\textbf{92.13} &50.31 &65.08
                  &70.79 &37.95 &49.41

                  &78.26 &32.73 &46.15
                  &52.17 &21.82 &30.77
                  \\
UTOS+Ada-TSA      &18.82 &21.47 &20.06
                  &10.75 &12.05 &11.36
                  
                  &22.06 &27.27 &24.39
                  &5.88 &7.27 &6.50

            \\
UTOS+DANN         &85.32 &57.06 &68.38
                  &60.91 &40.36 &48.55

                  &78.12 &45.45 &57.47
                  &37.50 &21.82 &27.59

            \\
UTOS+MEDM         &39.21 &54.60 &45.64
                  &6.45 &30.12 &10.63
                  
                  &46.67 &50.91 &48.70
                  &9.3 &21.82 &13.04

            \\
UECA-Prompt       &75.14 &81.60 &78.24
                  &66.27 &65.87 &66.07

                  &75.93 &74.55 &75.23
                  &58.18 &58.18 &58.18

                  \\
Ours              &80.72 &82.21 &81.46
                  &\textbf{84.71} &\textbf{79.64} &\textbf{82.10}
                  
                  &84.31 &78.18 &81.13
                  &\textbf{83.33} &\textbf{81.82} &\textbf{82.57}
                  \\
\noalign{\hrule height 1.8pt}
\end{tabular}
\end{adjustbox}
\caption{Experimental results of our models and baselines utilizing precision (P), recall (R), and F1 score (F1) as metrics on the \task task. Emotion Extraction is denoted by EE. S refers to source domain.}
\label{table: exp_results_large}
\end{table*}

\end{document}